\documentclass[10pt,twocolumn,letterpaper]{article}

\usepackage[pagenumbers]{cvpr} 

\definecolor{cvprblue}{rgb}{0.21,0.49,0.74}
\usepackage[pagebackref,breaklinks,colorlinks,allcolors=cvprblue]{hyperref}
\usepackage{multirow}

\def\eg{\emph{e.g}\onedot\xspace} 

\def\ie{\emph{i.e}\onedot\xspace}

\def\name{LiftImage3D\xspace}
\def\dustthreer{MASt3R\xspace} %

\definecolor{metric-1}{HTML}{ff9999}
\definecolor{metric-2}{HTML}{ffcc99}
\definecolor{metric-3}{HTML}{fff6b2}

\def\paperID{} 
\def\confName{CVPR}
\def\confYear{2025}

\title{LiftImage3D: Lifting Any Single Image to 3D Gaussians \\ with Video Generation Priors}

\author{Yabo Chen$^{1}$\footnotemark[1], \;\; Chen Yang$^{1}$\footnotemark[1], \;\; Jiemin Fang$^{2}$\textsuperscript{$\dagger$}, \;\; Xiaopeng Zhang$^{2}$, \\
\;\; Lingxi Xie$^{2}$, \;\; Wei Shen$^{1}$, \;\; Wenrui Dai$^{1}$, \;\; Hongkai Xiong$^{1}$, \;\; Qi Tian$^{2}$\\
$^1$Shanghai Jiao Tong University  \;\; $^2$Huawei Inc.\\ 
\tt\small{\{chenyabo, ycyangchen, wei.shen, daiwenrui, xionghongkai\}@sjtu.edu.cn}\\
\texttt{\small\{jaminfong, 198808xc, zxphistory\}@gmail.com} \;\;
\texttt{\small tian.qi1@huawei.com}\\
\url{liftimage3d.github.io}
}

\begin{document}
\twocolumn[{%
\renewcommand\twocolumn[1][]{#1}%
\maketitle
\begin{center}
\centering
\includegraphics[width=1.0\linewidth]{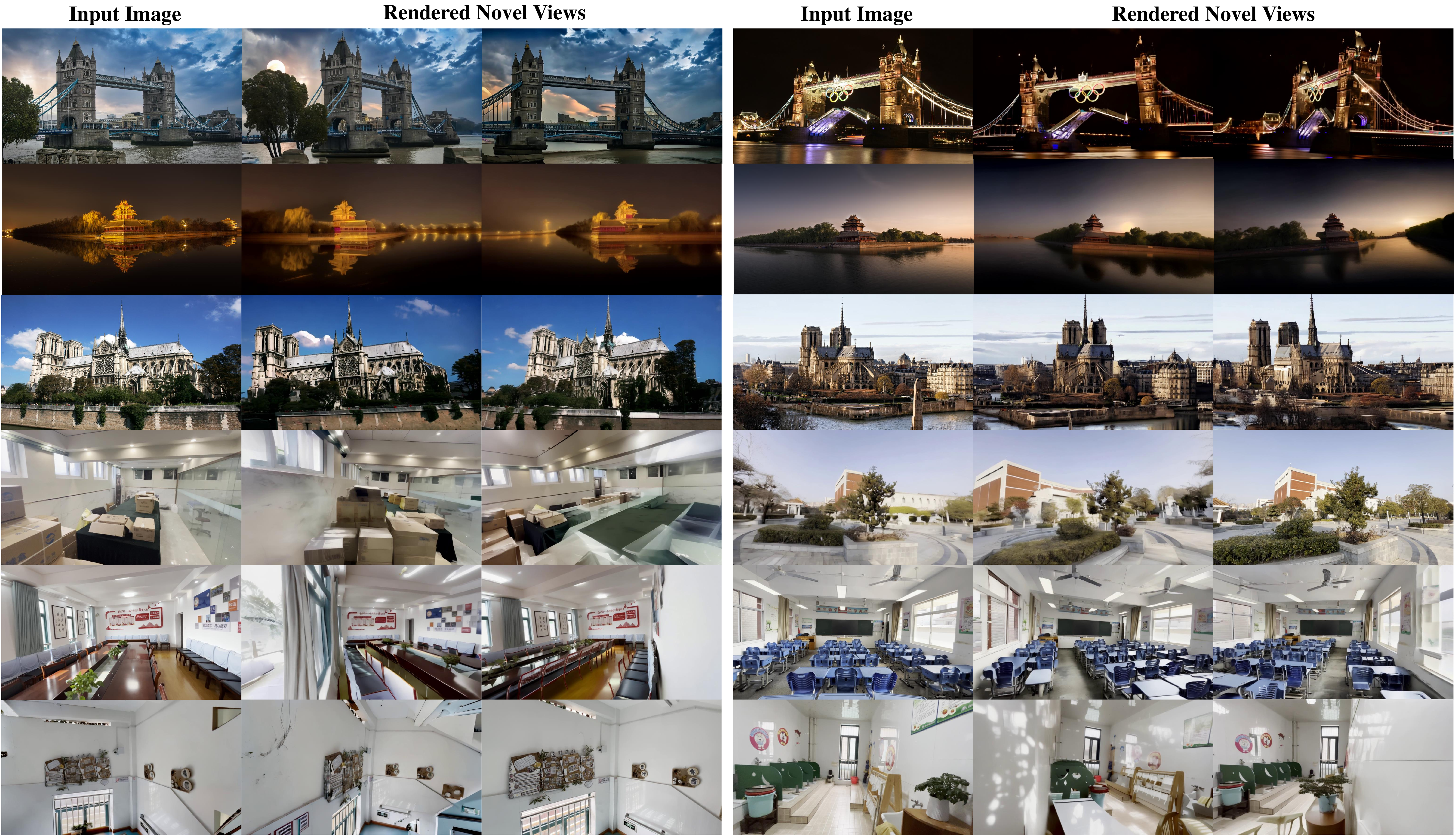}
\vspace{-20pt}
\captionof{figure}{LiftImage3D is a universal framework that utilizes video generation priors to lift any single 2D image into 3D Gaussians, capable of handling in-the-wild 3D objects/scenes.
}
\label{fig: teaser}
\end{center}
}]

{
\renewcommand{\thefootnote}{\fnsymbol{footnote}}
\footnotetext[1]{Equal contributions in no particular order.}
\footnotetext[2]{Project lead.}
}

\begin{abstract}
Single-image 3D reconstruction remains a fundamental challenge in computer vision due to inherent geometric ambiguities and limited viewpoint information. Recent advances in Latent Video Diffusion Models (LVDMs) offer promising 3D priors learned from large-scale video data. However, leveraging these priors effectively faces three key challenges: (1) degradation in quality across large camera motions, (2) difficulties in achieving precise camera control, and (3) geometric distortions inherent to the diffusion process that damage 3D consistency. We address these challenges by proposing LiftImage3D, a framework that effectively releases LVDMs' generative priors while ensuring 3D consistency. 
Specifically, we design an articulated trajectory strategy to generate video frames, which decomposes video sequences with large camera motions into ones with controllable small motions. Then we use robust neural matching models, i.e. MASt3R, to calibrate the camera poses of generated frames and produce corresponding point clouds. Finally, we propose a distortion-aware 3D Gaussian splatting representation, which can learn independent distortions between frames and output undistorted canonical Gaussians. 
Extensive experiments demonstrate that LiftImage3D achieves state-of-the-art performance on two challenging datasets, i.e. LLFF, DL3DV, and Tanks and Temples, and generalizes well to diverse in-the-wild images, from cartoon illustrations to complex real-world scenes.
\end{abstract}    
\section{Introduction}
\label{sec:intro}
Lifting a single image into a realistic 3D scene is a crucial step towards enabling immersive media and interactive experiences in applications like virtual/augmented reality content creation. However, this task remains challenging due to the inherent geometry ambiguity and extremely sparse information available from a single view. While traditional methods rely on multiplane image (MPI), depth warping and image inpainting~\cite{Adampi, SinMPI, luciddreamer, Xu_2022_SinNeRF,barron2023tiled3dphotography}, recent advances in Latent Video Diffusion Models (LVDM) offer a promising new direction. These models have demonstrated remarkable capabilities in generating temporally coherent videos with rich implicit 3D understanding learned from large-scale video data, suggesting their potential as powerful priors for single-image 3D reconstruction.

Integrating rich 3D priors from LVDMs into single image reconstruction but in a controllable manner
is a non-trivial task with three main challenges:
1) \emph{Collapse with Large Camera Motions}. While existing methods can achieve high-quality novel view synthesis for small camera motions, their rendering quality degrades dramatically as the camera moves drastically. This limitation arises with the accumulation of generation errors across large viewpoint changes. The difficulty in maintaining geometric consistency across distant views severely restricts the range of achievable novel views.
2) \emph{Inaccurate Camera Control}. Although LVDMs excel at generating temporal-coherent videos, controlling their generation to follow desired camera trajectories remains challenging. 
Recent attempts with significant limitations~\cite{voleti2024sv3d, Han2024vfusion3d, melas2024im3d,kwak2023vivid123, pang2024envision3d, wang2023motionctrl} explicitly incorporates camera trajectories as LVDM conditions through fine-tuning or re-training. However, the lack of underlying 3D information of the scene results in a mismatch between the generated results and the input camera condition.
3) \emph{3D-Inconsistent Distortions}. The inherent noising and de-noising procedure of diffusion models introduces subtle yet persistent geometric distortions during frame generation. These distortions, while imperceptible visually, accumulate across views and severely damage 3D reconstruction quality by corrupting geometric consistency between frames.

To address these challenges, we propose LiftImage3D, a novel framework that can lift any single image to high-quality 3D scenes by effectively harnessing LVDM's generative priors while ensuring 3D consistency. 
First, we design an articulated trajectory strategy that decomposes large camera motions into small controllable steps. The generation quality can be guaranteed while achieving wider view coverage to overcome limited camera motion.
For inaccurate camera control, we directly estimate camera trajectories and coarse geometries from generated frames using robust neural matching methods~\cite{dust3r_cvpr24, mast3r, monst3r}.  
Direct pose estimation bypasses the impact of inconsistency between the input condition and generated frames.
To handle 3D-inconsistent distortions, we design a distortion-aware 3D Gaussian splatting (3DGS)~\cite{kerbl3Dgaussians}, which consists of five-dimensional HexPlanes~\cite{cao2023hexplane}, combining xyz spatial coordinates (canonical space) with a two-dimensional time axes stamps. 
In this way, we extract 3D consistent priors from LVDMs and use these priors to build the canonical space while generation-induced distortions are explicitly modeled via deformation fields. 
Furthermore, we leverage the coarse geometry from neural matching to calibrate monocular depth estimates with proper scales and shifts, providing accurate and 3D-consistent depth priors.

Our contributions can be summarized as follows:
\begin{itemize}[leftmargin=10pt]
\item  We have proposed a systematic framework that can lift any single image to 3D Gaussians, releasing the 3D capabilities of latent video diffusion models (LVDM) but in a \textbf{controllable} manner.
\item We present an articulated trajectory strategy, a frames matching strategy, a depth prior injection, and an effective representation \ie~distortion-aware 3DGS to solve the challenges in reconstruction from generated video frames to 3D.
\item Extensive experiments on three widely used datasets, LLFF~\cite{mildenhall2019llff}, DL3DV~\cite{dl3dv}, and Tanks and Temples~\cite{tanksandtemples}, demonstrate that LiftImage3D achieves higher visual quality and better 3D consistency compared to previous SOTAs. 
It also performs well on in-the-wild images. As shown in Figure~\ref{fig: teaser}., LiftImage3D generalizes well to diverse inputs, ranging from cartoon-like illustrations to complex real-world scenes, including both indoors and outdoors ones.
\end{itemize}
\section{Related Work}
\label{sec:relatedworks}

\subsection{3D Photography}
The term 3D photography refers to the use of novel view synthesis in the wild scenes from a single image~\cite{barron2023tiled3dphotography}. These layer-based methods tend to use MPI (multiplane image)~\cite{szeliski1998,zhou2018,srinivasan2019,wizadwongsa2021}, which can produce synthesis results from a single input image by using multiple planes per pixel
to represent a scene. Many MPI-based methods~\cite{li2020,Adampi,luvizon2021} are built to build 3D photography. 
Another efficient approach to 3D photography involves depth-based warping~\cite{chaurasia2020,xiao2022}. These methods 
usually guide inpainting in occluded regions of warped views.~\cite{shih20,li2022,wiles2020,niklaus2019,choi2019}
Nonetheless, depth-based warping suffers from
hard boundaries and is over-sensitive to errors in the depth
estimate~\cite{khakhulin2022}.
Some methods have tried to use text-to-image diffusion methods to generate out-of-view contents and project all scene contents into an expanded multiplane image according to depths~\cite{guo2023SinMPI,xiang20233daware} to incorporate information beyond the input image.
However, even incorporating latent diffusion model(LDM) into MPI, these methods provide an expanded novel view prior that is still based on planes. The generated expanded viewpoints may exhibit a significant flatness due to the nature of the image-based latent diffusion approach. Additionally, repeated usage of the image-based LDM can lead to a decrease in the obtained 3D consistency.

\subsection{2D Generated Model Based Single Image to 3D}
Many researchers have studied the tasks of generating 3D models and achieving novel view synthesis using only a single image~\cite{woo2023harmonyview,simon2024hypervoltran,zhang2023repaint123,kocsis2023intrinsic,li2023valid,wu2023hyperdreamer,liu2023one2345,yu2023boosting3d,chen2023singleview,shi2023tosshighquality}. Some researchers directly train a 3D model on 3D data~\cite{chan2023genvs,xiang20233daware,lee2023localaware,nichol2022point,jun2023shapee,qian2024pushing,huang2024zeroshape}, but they tend to have good generation quality only on scenes similar to the training set.
Recently, by constructing a conditional latent diffusion model(LDM) based on camera viewpoints, many works have made it possible to pre-train a single image-to-3D model base on text-to-image latent diffusion models.~\cite{liu2023zero1to3,shen2023anything,tang2023make,xu2023neurallift,liu2023one,melas2023realfusion,qian2023magic123,hamdi2023sparf,shi2023zero123++,lin2023consistent123,shi2023toss,sargent2023zeronvs,liu2023syncdreamer,long2023wonder3d,ye2023consistent,weng2023zeroavatar,hu2023humanliff,yang2023consistnet,weng2023consistent123,tang2024mvdiffusion}. These methods learn from large-scale multi-view images~\cite{deitke2023objaverse,objaverseXL} to build the geometric priors of large-scale diffusion models.
However, these models are deprived of their ability to generate scenes. Because they mainly focus on object-level single image to 3D and lack pre-training on large-scale scene datasets.
Some other researchers~\cite{luciddreamer, wonderworld} borrow the ability of diffusion based inpainting models and build the problem of single image to 3D scene as a inpainting problem. These methods predict depth from input images, then back-project to obtain corresponding 3D representations. Subsequently, based on the prior information of depth, extrapolation inpainting is performed to complete more 3D priors. However, these methods often rely on relatively precise text prompts. The final quality and consistency of the 3D representation usually can only be maintained in relation to the text prompt, making it challenging to ensure consistency with the input images.

\subsection{LVDM-Based Single Image to 3D}
Many researchers believe that video diffusion models can provide a strong multi-view 3D-prior.~\cite{Blattmann2023svd, li2024sora, liu2024sora}. However, generated videos do not contain camera poses, making it challenging to integrate them directly with existing 3D representations.
Recent researches try to finetune a multi-view diffusion model from the LVDM on multi-view datasets.~\cite{voleti2024sv3d, Han2024vfusion3d, melas2024im3d,kwak2023vivid123, pang2024envision3d, viewcrafter}. However, while the diffusion model enhances the ability to synthesize novel views of objects, it also results in the loss of scene and real-world perception from the latent video diffusion model. 

MotionCtrl~\cite{wang2023motionctrl}, NVS-Solver~\cite{nvssolver}, and View-Crafter~\cite{viewcrafter}, utilize LVDM to generate and model the novel view synthesis problems as video frames generation. However, directly use video frames to build a 3D representation may faced with many problems as discussed in the introduction. Including but not limited to collapse with large camera motions, inaccurate generation with controlled camera, and 3D inconsistent distortions.

\begin{figure*}
    \centering
    \includegraphics[width=\linewidth]{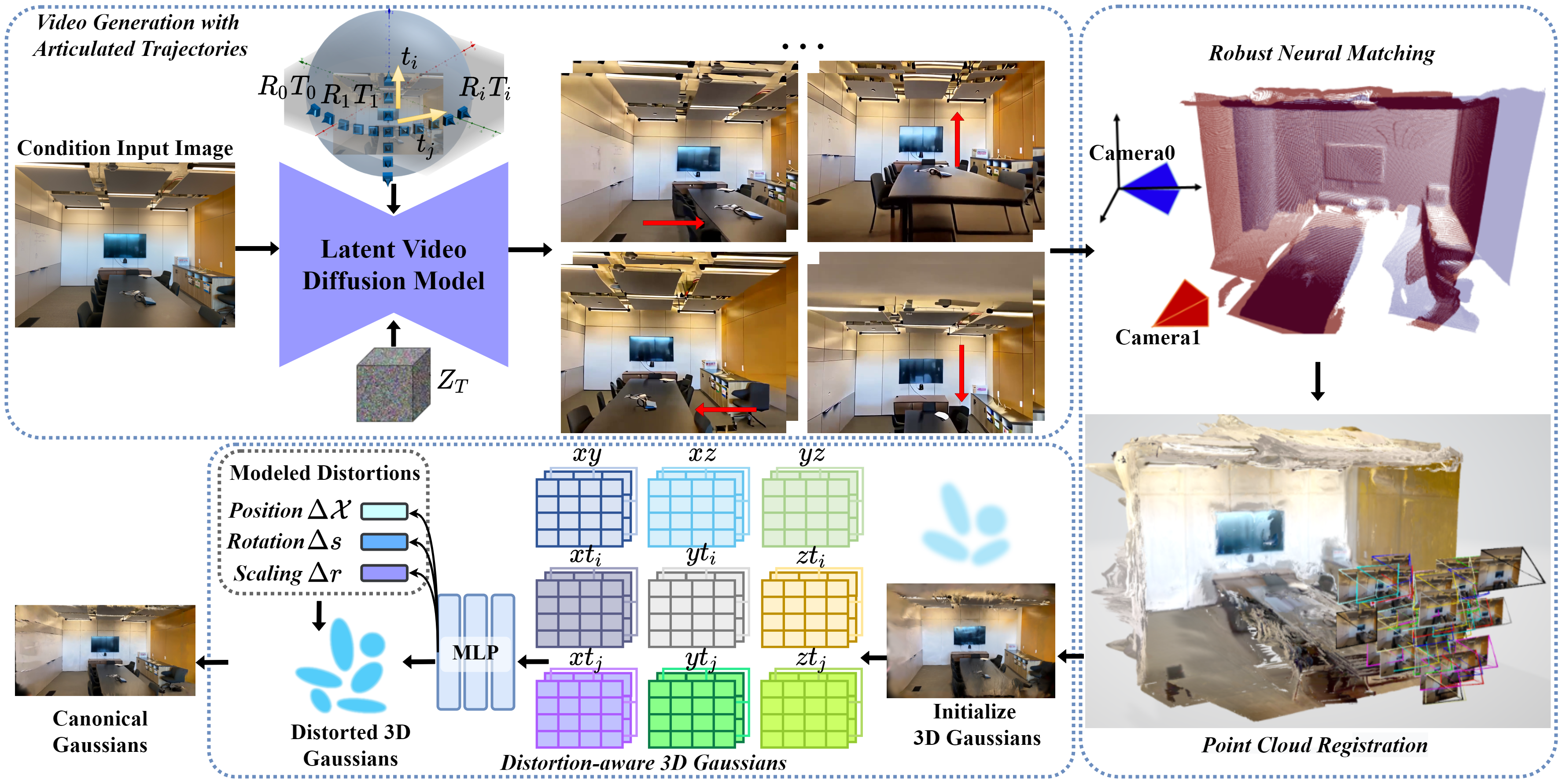}
    \vspace{-15pt}
    \caption{
    The overall pipeline of \name. We first extend LVDM to generate multiple video clips from a single image using an articulated camera trajectory strategy. 
    Then all generated frames are matched using the robust neural matching module and registered into a point cloud. After that, we initialize Gaussians from registered point clouds and construct a distortion field to model the independent distortion of each video frame upon canonical 3DGS. 
    }
    \label{fig:pipe}
    \vspace{-16pt}
\end{figure*}

\section{Methods}
\subsection{Preliminaries}
\textbf{3D Gaussian Splatting}
3D Gaussian Splatting~\cite{kerbl3Dgaussians} represents 3D scenes with 3D Gaussians, which is similar with the form of point clouds. Each 3D Gaussian is characterized by the center location $\mathcal{X}$, scaling vector $s$, rotation quaternion $r$, opacity $\sigma$, and spherical harmonic (SH) coefficients $\mathcal{SH}$. Thus, a scene is parameterized as a set of Gaussians $\mathcal{G}= \left\{\mathcal{X}, s, r, \sigma, \mathcal{SH}\right\}$.

Many researches~\cite{yang2024gaussianobject, ye2024noposplat} have shown its robust general reconstruction capabilities and can easily integrate into the point cloud representation.
\paragraph{Motion-controllable LVDM}
Stable Video Diffusion (SVD)~\cite{Blattmann2023svd} is a representative method of the LVDM. SVD can generate high-quality videos guided by a single image prompt. It employs a denoising diffusion model $f_v$ in the latent space for computational and memory efficiency. Motion-controllable LVDM~\cite{wang2023motionctrl, viewcrafter} extends SVD by introducing camera motion control into the LVDM generation process. Since camera motions represent global transformations between video frames. Specifically, MotionCtrl~\cite{wang2023motionctrl} incorporates a Camera Motion Control Module (CMCM) that interfaces with SVD through its temporal transformers. The CMCM takes a sequence of camera poses as input:
\begin{equation}
    RT = \{R_1T_1, R_2T_2, \ldots, R_{l}T_{l}\},
\end{equation}
where $RT \in \mathbb{R}^{l \times 3 \times 4}$ and $l$ denotes the video length. 
With CMCM, the frame generation of LVDM $f_v$ can be formulated as:
\begin{equation}
    x_i^{\prime} = f_v(x, R_iT_i),\quad i \sim [0, 1,...,L],
\end{equation}
where $x_i^{\prime}$ is the $i$-th generated frame; $x$ is the input image.

\paragraph{Learning-based Visual Odometry} 
The learning-based visual odometry model~\cite{dust3r_cvpr24, mast3r, monst3r} aims to integrate the SfM (Structure from Motion) and MVS (Multi-View Stereo) processes together, which estimates point maps $\boldsymbol{P}_i \in \mathbb{R}^{H \times W \times 3}| i \sim [0, 1,...,L]$ directly from images with uncalibrated/unposed cameras poses $R^{\prime} T^{\prime} \in\mathbb{R}^{H \times W \times 3 \times 4}| i \sim [0, 1,...,L]$.

Specifically, the recent SOTA visual odometry model \dustthreer~\cite{mast3r}, is based on regression of the unprojected and normalized point maps of two input views. 
First, \dustthreer pairs all input images in groups of two. For each pair contains view1 and view2, through depth back-projection, we can obtain point maps of $\hat{\boldsymbol{P}}_{1,1}$ and $\hat{\boldsymbol{P}}_{2,1}$. The camera origin is set as view1. In this context, the subscript $\{2,1\}$ in $\hat{\boldsymbol{P}}_{2,1}$ signifies that the origin of the coordinate system for view2 is anchored at view1.

Then the relative pose estimation can be achieved in a direct way which aligns the pointmaps $X^{1,1} \leftrightarrow X^{1,2}$ (or, equivalently, $X^{2,2} \leftrightarrow X^{1,2}$) using Procrustes alignment~\cite{Procrustesalignment} to get the relative pose $P^*=\left[R^* \mid T^*\right]$ :
\begin{equation}
\begin{aligned}
R^*, T^*=\underset{\sigma, R, T}{\arg \min } \sum_i C_i^{1,1} C_i^{1,2}\left\|\sigma\left(R X_i^{1,1}+T\right)-X_i^{1,2}\right\|^2.
\end{aligned}
\end{equation}

\subsection{Overall Framework}
The core of \name is to create a pipeline that extends the capabilities of the video generation model to the task of transforming a single image into a 3D scene representation in a controllable manner.
To achieve it, as shown in Figure~\ref{fig:pipe}., our \name framework consists of three main components: video frame generation, camera pose neural matching and point cloud registration, and distortion-aware 3DGS optimization.
In Section~\ref{subsec:latentdiffusion}, we extend LVDM to generate diverse video clips from a single image using an articulated camera trajectory strategy. Section~\ref{subsec:dust3r} introduces our efficient robust neural matching module, which employs temporal consistency priors in a matching graph framework to minimize computational costs during point cloud registration. 
Section~\ref{subsec:gaussian_training} presents our distortion-aware 3DGS optimization pipeline, where we initialize Gaussians from registered point clouds and construct a distortion field to model the independent distortion of each video frame upon canonical 3DGS. Here, canonical 3DGS refers to the accurate undistorted 3D scene representation according to the input image. In this section, we introduce the network structure of distortion modeling, depth prior injection, and loss function to make the distortion field fit the input image. 

\begin{figure}
    \centering
    \includegraphics[width=\linewidth]{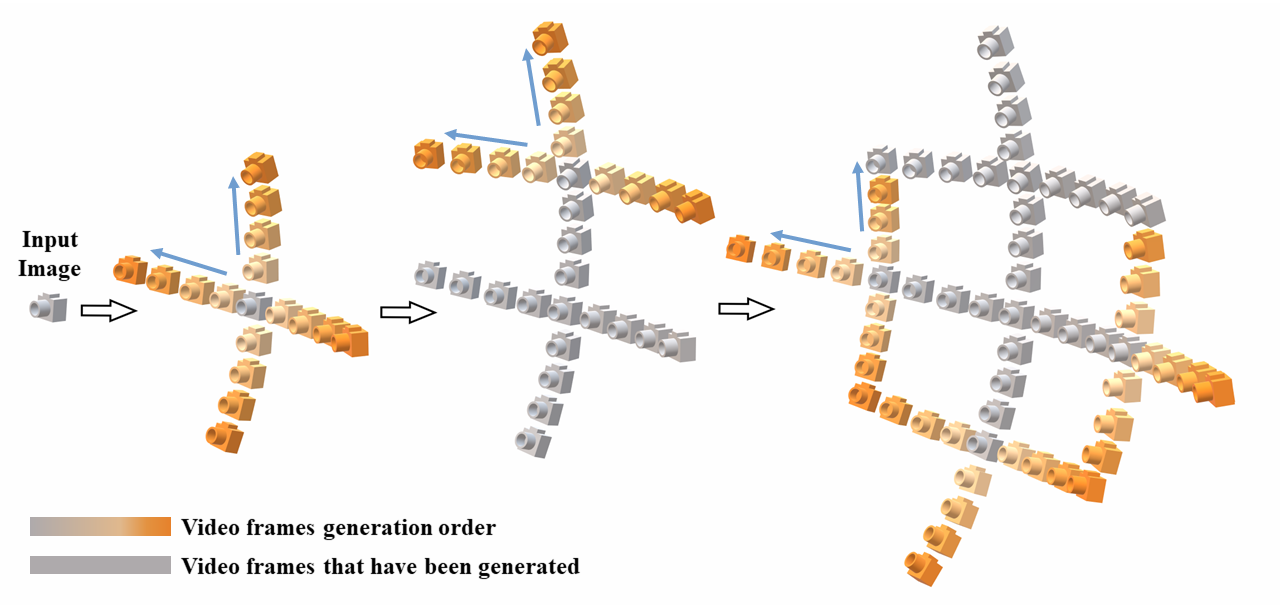}
    \vspace{-14pt}
    \caption{Articulated trajectory generation pipeline. The gray cameras indicate previously generated frames, and the orange cameras show the current generation sequence.
    The process iteratively generates frames following a predefined trajectory (orange gradient), where each subsequent generation uses the terminal frame from the previous sequence as its input. This cascading approach enables comprehensive object coverage through controlled camera trajectories.
    }
    \label{fig:articulated}
     \vspace{-10pt}
\end{figure}

\subsection{Video Generation with Articulated Trajectories}
\label{subsec:latentdiffusion}
Current motion-controllable LVDMs struggle to generate high-quality videos with significant camera motions, often producing frames with visible distortions and blurs when the camera moves drastically. To achieve wider view coverage while maintaining frame quality, we introduce an articulated generation strategy that decomposes large camera motions into smaller, controllable steps. 

We propose to generate video frames following articulated trajectories, where generated frames are used as new starting anchors for subsequent video generation. This strategy enables us to achieve larger view ranges while maintaining frame quality by limiting each generation step to small, stable camera motions. 

Formally, we define the camera moves towards $D$ directions, then the camera trajectories can be defined as a set of poses $\{R_i T_i~|~i \sim [0, ..., l \times D]\}$, where $l$ denotes the length of the generated video.
Then we can generate views from the input image as $x_i^{\prime}=f_v\left(x, R_i T_i\right) | i \sim [0, 1,...,l \times D]$, where $f_v$ is the LVDM networks, and $i$ is the generated frame index. These frames can be generated with smooth motion movement. Then we use these views $x_{d_i}^{\prime}=f_v\left(x, R_D T_D\right) | d_i \sim [0, ..., D-1]$ as a new anchor to generate new video frames articulately.

The generated frames from LVDM $f_v$ using the articulated strategy can be formulated as:
\begin{small}
\begin{equation}
\begin{aligned}
\begin{cases}
x_i^{\prime}=f_v\left(x, R_i T_i\right) | i \sim [0, 1,..., l \times D] \\
x_i^{\prime\prime}=f_v\left(x_{d_i}^{\prime}, R_i T_i\right) | i \sim [0, ..., l \times (D-1)], d_i \sim [0, ..., D]\end{cases}
\end{aligned}  
\end{equation}
\end{small}The second generation stage only has $D-1$ potential camera pose moving directions, as there is no need to have a trajectory to move back to the input image. Totally, taking the example of accumulating twice, we have $L = l \times D + (l-1) \times (D-1)$ frames. In practice, frames are mainly generated along the directions of two axes (\ie~D = 4) as shown in Figure~\ref{fig:articulated}.

\subsection{Robust Neural Matching}
\label{subsec:dust3r}
The generation process of LVDM involves temporal noises from the denoising process, resulting in inherent generation randomness. The nosing and denoising procedure produces geometric distortions and lacks strict 3D consistency across frames, as the generation process does not explicitly enforce multi-view geometric constraints. These fundamental limitations make traditional structure-from-motion (SfM) methods, \eg COLMAP~\cite{schoenberger2016sfm, schoenberger2016mvs}, ineffective for pose estimation. 

To address this challenge, we adopt a dense neural matching method, \dustthreer~\cite{mast3r}, to provide robust pose estimation and build point clouds from these generated frames. 
However, \dustthreer processes all the images in pairs, forming the registered point clouds and estimating the camera pose by matching the pointmaps of any two images. For L sparse frames, \dustthreer basically needs to build $L \times L-1$ pairs for precise projection, which is cumbersome. The order of single image to video frames determines the temporal priors between video frames. We simplify the pairs following the temporal priors. Each image is only paired with the frame generated just before it in the sequence, resulting in only $L-1$ pairs

In detail, the first step of neural matching is to recover camera intrinsics using the Weiszfeld algorithm~\cite{Weiszfeldalgorithm}. 
Since all frames are generated from the same input image, we assume the frames share the camera intrinsics, which simplifies the modeling process. 
Specifically, we compute the optimal focal length $f^*$ by minimizing:
\begin{equation}
f^* = \underset{f}{\arg \min } \sum_{i=0}^W \sum_{j=0}^H \boldsymbol{O}^{i, j}\left\|\left(i^{\prime}, j^{\prime}\right)-f \frac{\left(\boldsymbol{P}^{i, j, 0}, \boldsymbol{P}^{i, j, 1}\right)}{\boldsymbol{P}^{i, j, 2}}\right\|,
\end{equation}
where $(i^{\prime}, j^{\prime}) = (i-\frac{W}{2}, j-\frac{H}{2})$ represents centered pixel coordinates with respect to the image center. The optimized focal length $f^*$ is then used across all subsequent procedures.

After that, we group all video frames into pairs following the temporal priors, in which we pair each image in the sequence with the frame before it in a one-by-one manner. 
For each pair, comparing the pointmaps $X^{R_1T_1,R_1T_1} \leftrightarrow X^{R_1T_1,R_2T_2}$ with Procrustes alignment~\cite{Procrustesalignment}, we can get the precise relative pose $P^*=\left[R^* \mid T^*\right]$ as follows:
\begin{equation}
\begin{aligned}
R_1^*, T_1^*=\underset{\sigma, R, T}{\arg \min } \sum_i C_i^{R_1T_1,R_1T_1}  C_i^{R_1T_1,R_2T_2} \\
\left\|\sigma\left(R X_i^{R_1T_1,R_1T_1}+T\right)-X_i^{R_1T_1,R_2T_2}\right\|^2,
\end{aligned}
\end{equation}
where $C$ is the rendering function. In summary, we can get camera poses $R_i^* T_i^* | i \sim [0, 1,...,L]$ exactly from generated video frames using a set of simple and fast regressions. 

After obtaining the camera pose and intrinsic parameters, as well as the point cloud of each video frame, all the point clouds can be merged, referred to as the registered point clouds.

\subsection{Distortion-aware 3D Gaussian Splatting}
\label{subsec:gaussian_training}
\paragraph{Distortion-aware Network Structure}
3DGS~\cite{kerbl3Dgaussians} has robust general reconstruction capabilities. To convert the previously registered point clouds into a higher-quality 3D representation for rendering, we utilize 3DGS, which can be seamlessly integrated with point cloud representations.
However, the noising and denoising process of diffusion models introduces various geometric distortions during frame generation. These distortions severely corrupt the geometric consistency of 3DGS, resulting in artifacts, \eg blurred details, geometric distortions, and edge fringing \emph{etc}. We need to extract and detach these distortions to construct canonical 3D Gaussians $\mathcal{G}= \left\{\mathcal{X}, s, r, \sigma, \mathcal{SH}\right\}$, which model the point cloud location $\mathcal{X}$, scaling vector $s$, rotation quaternion $r$, opacity $\sigma$, and spherical harmonic (SH) coefficients $\mathcal{SH}$. 
The canonical 3DGS refers to the accurate undistorted 3D scene representation corresponding to the precise undistorted input image.

Inspired by 3DGS methods with deformation fields~\cite{wu20234dgaussians, yang2023deformable3dgs}, we propose a distortion-aware 3DGS, which contains a distortion field to represent the canonical-to-distortion mapping.
\begin{equation}
\begin{aligned}
\mathcal{F}:(\mathcal{G} + \Delta \mathcal{G}, \boldsymbol{t}) \rightarrow \mathcal{G}
\end{aligned}
\end{equation}
where $\mathcal{G}$ is canonical 3DGS and the distortion $\Delta \mathcal{G}$ is modeled by the distortion field network $\mathcal{F}$. The main task of $\mathcal{F}$ is to model the independent
distortion $\Delta \mathcal{G}$ of each video frame and extract the canonical 3DGS $\mathcal{G}$.
As the frames are generated along two axes, a two-dimensional stamp $\boldsymbol{t} = \{\boldsymbol{t}_i, \boldsymbol{t}_j\}$ is assigned to each frame to differentiate these frames and the corresponding temporal distortion. Specifically, we model the two-axis stamp $\boldsymbol{t}$ along the left-right (x-direction) as $\boldsymbol{t}_i$ and up-down (y-direction) as $\boldsymbol{t}_j$ for the input image.

Specifically, inspired by ~\cite{wu20234dgaussians, cao2023hexplane, kplanes_2023}, the distortion field network $\mathcal{F}$ consists of three parts. A spatial-temporal structure encoder $\mathcal{H}$ with a multi-resolution HexPlane~\cite{cao2023hexplane} $R(n_i, n_j)$, combine xyz spatial coordinates (canonical space) with a two-dimensional time axes stamps. A tiny MLP $\phi_d$ merges all the features.

Because modeling the three-dimensional xyz spatial and two-dimensional time axes stamps with the vanilla 5D neural voxels are memory-consuming, we adopt a 5D K-Planes~\cite{kplanes_2023} module to decompose the 5D neural voxel into 9 multi-resolution planes. 
The spatial-temporal structure encoder $\mathcal{H}$ contains 9 multi-resolution plane modules $R_l(n_i, n_j)$, ~\ie the K-Planes of $(x, y), (x, z), (y, z), (x, \boldsymbol{t}_i), (y, \boldsymbol{t}_i), (z, \boldsymbol{t}_i), (x, \boldsymbol{t}_j), (y, \boldsymbol{t}_j), (z, \boldsymbol{t}_j)$. $\boldsymbol{t}_i$ and $\boldsymbol{t}_j$ are constructing the same trajectory space, so we do not need to construct the K-Planes of $\boldsymbol{t}_i\boldsymbol{t}_j$. 
Each voxel module is defined by $R(n_i, n_j) \in \mathbb{R}^{h \times N_l N_i \times N_l N_j}$, where $h$ stands for the hidden dimension of features, and $N_i$,$N_j$ denotes the basic resolution of voxel grid and we use a multi-resolution structure with an upsampling scale $N_l$. 
The formula for computing separate voxel features is as follows:
\begin{equation}
\begin{aligned}
f_h = \bigcup_l &\prod \operatorname{interp}\left(R_l(n_i, n_j)\right), \; (n_i, n_j)\\
\in \{&(x, y), (x, z), (y, z), (x, \boldsymbol{t}_i), (y, \boldsymbol{t}_i), \\
&(z, \boldsymbol{t}_i), (x, \boldsymbol{t}_j), (y, \boldsymbol{t}_j), (z, \boldsymbol{t}_j)\}&
\end{aligned}
\end{equation}
$f_h \in \mathbb{R}^{h * N_l}$ is the feature of neural voxels. `interp' denotes the bilinear interpolation for querying the voxel features at 5 vertices of the grid. Then a tiny MLP $\phi_d$ merges all the features by $f_d=\phi_d\left(f_h\right)$.

For each 3DGS, the distortion can be modeled in position $\Delta \mathcal{X}=$ $\phi_x\left(f_d\right)$, rotation $\Delta r=\phi_r\left(f_d\right)$, and scaling $\Delta s=\phi_s\left(f_d\right)$ separately, where $f_d$ is the encoded features of the 3DGS. The distorted feature $\left(\mathcal{X}^{\prime}, r^{\prime}, s^{\prime}\right)$ can be computed as: 
\begin{equation}
\begin{aligned}
\left(\mathcal{X}^{\prime}, r^{\prime}, s^{\prime}\right)=(\mathcal{X}+\Delta \mathcal{X}, r+\Delta r, s+\Delta s)
\end{aligned}
\end{equation}


When all the features of 3DGS are encoded, distortion-aware 3DGS will use a multi-head Gaussian distortion decoder $\mathcal{D}=\left\{\phi_x, \phi_r, \phi_s\right\}$ to model the distorted 3DGS. 
After training the 3DGS, these modeled distortions $\Delta \mathcal{X}$, $\Delta r$, and $\Delta s$ are discarded and only the canonical 3DGS are reserved.

\paragraph{Depth Prior Injection}
\label{subsec:depthprior}
To produce smoother and depth-consistent results, we leverage the coarse but absolute depth maps from neural matching to calibrate the fine depth map from monocular depth estimation with proper scales and shifts. In practice, we employed the recent state-of-the-art monocular depth estimation method, ~\ie Depth Anything V2~\cite{depth_anything_v2}.
Depth prior injection can provide more accurate and 3D-consistent depth results.

The depth maps produced by neural matching are coarse but have absolute depth. Each pixel value directly corresponds to a physical distance. The depth maps produced by monocular depth estimation are finely detailed but have no metric distance.
Therefore, we calibrate the two distributions of coarse neural matching depth (absolute depth) $d_a$ and fine monocular depth (relative depth) $d_r$ as follows:
\begin{equation}
\begin{aligned}
\operatorname{Scale} &= \operatorname{med}(\hat{d_a}/\hat{d_r}),\\
\operatorname{Shift} &= \operatorname{med}(\hat{d_a} - \operatorname{Scale} \cdot \hat{d_r}), \\
d &= \operatorname{Scale} \cdot d_r + \operatorname{Shift},
\end{aligned}
\end{equation}
where $\operatorname{med}$ represents the median value. $\hat{d_a}$ and $\hat{d_r}$ are the centered depths defined as $\hat{d_a} = d_a - \operatorname{med}(d_a)$ and $\hat{d_r} = d_r - \operatorname{med}(d_r)$, respectively.
Then, we can use $d$ as the calibrated depth prior to supervising the acquisition of a higher-quality 3D representation with improved depth quality. 

\paragraph{Loss Function Design}
In the optimizing process of the distortion field, we need to ensure that the modeling distortion does not deviate too far from the canonical 3DGS. Completely pursuing modeling the distortion can result in poor canonical 3DGS. Therefore, we design a loss function that models around the precise undistorted input image, where the distortion $\Delta \mathcal{X}$, $\Delta r$, and $\Delta s$ should be minimized as much as possible in that vicinity. 
\begin{small}
\begin{equation}
\begin{aligned}
\mathcal{L}_{distort} = |\Delta \mathcal{X} - 0|_{\boldsymbol{t}_i\boldsymbol{t}_j=0} + |\Delta r - 0|_{\boldsymbol{t}_i\boldsymbol{t}_j=0} + |\Delta s - 0|_{\boldsymbol{t}_i\boldsymbol{t}_j=0}.
\end{aligned}
\end{equation}
\end{small}

For the training process, we use the $\mathcal{L}1$ RGB loss and an LPIPS loss~\cite{lpipsloss} $\mathcal{L}_{lpips}$ to supervise the similarity of texture and reconstruction. A grid-based total-variational loss~\cite{niemeyer2021regnerf} $\mathcal{L}_{tv}$ is also applied. To prevent distortion from growing disorderly.
Another $\mathcal{L}1$ depth loss is built upon the calibrated depth priors $d$ and rendering depths.

\begin{table*}
\caption{Quantitative comparison of our method against other state-of-the-art methods evaluated on the LLFF, DL3DV and Tanks and Temples datasets.}
  \label{sample-table}
  \centering
  \small
    \begin{tabular}{c|ccc|ccc|ccc}
    \hline 
    & \multicolumn{3}{c|}{LLFF dataset} & \multicolumn{3}{c}{DL3DV dataset} & \multicolumn{3}{c}{Tanks and Temples dataset}\\
    Method & PSNR $\uparrow$ & SSIM $\uparrow$ & LPIPS $\downarrow$ & PSNR $\uparrow$ & SSIM $\uparrow$ & LPIPS $\downarrow$ & PSNR $\uparrow$ & SSIM $\uparrow$ & LPIPS $\downarrow$ \\
    \hline 
    AdaMPI~\cite{Adampi} & 12.30 & 0.316 & 0.651 & 13.95 & 0.546 & 0.661 &11.36 & 0.457& 0.623 \\	
    SinMPI~\cite{SinMPI} & 12.89 & 0.309 & 0.602 & 14.35 & 0.531 & 0.611 & 11.52 & 0.490 & 0.591 \\
    LucidDreamer~\cite{luciddreamer} & 12.29 & 0.292 & 0.585 & 12.77 & 0.469 & 0.563 &12.95 & 0.394& 0.571 \\
    ViewCrafter~\cite{viewcrafter} & 13.51 & 0.296 & 0.577 & 17.16 & 0.608 & \textbf{0.407} &13.33& 0.429 &0.568 \\
    LiftImg3D (Ours) & \textbf{18.24} & \textbf{0.488} & \textbf{0.519} & \textbf{21.08} & \textbf{0.685} & 0.451 &\textbf{16.01} & \textbf{0.512} & \textbf{0.553} \\
    \hline
    \end{tabular}
\end{table*}

So the total loss can be formulated as follows:
\begin{equation}
\begin{aligned}
\mathcal{L} = \mathcal{L}1_{RGB} + \mathcal{L}_{lpips} + \mathcal{L}1_{depth} + \mathcal{L}_{tv} + \mathcal{L}_{distort}.
\end{aligned}
\end{equation}

\begin{figure*}[t]
    \centering
    \includegraphics[width=1.0\linewidth]{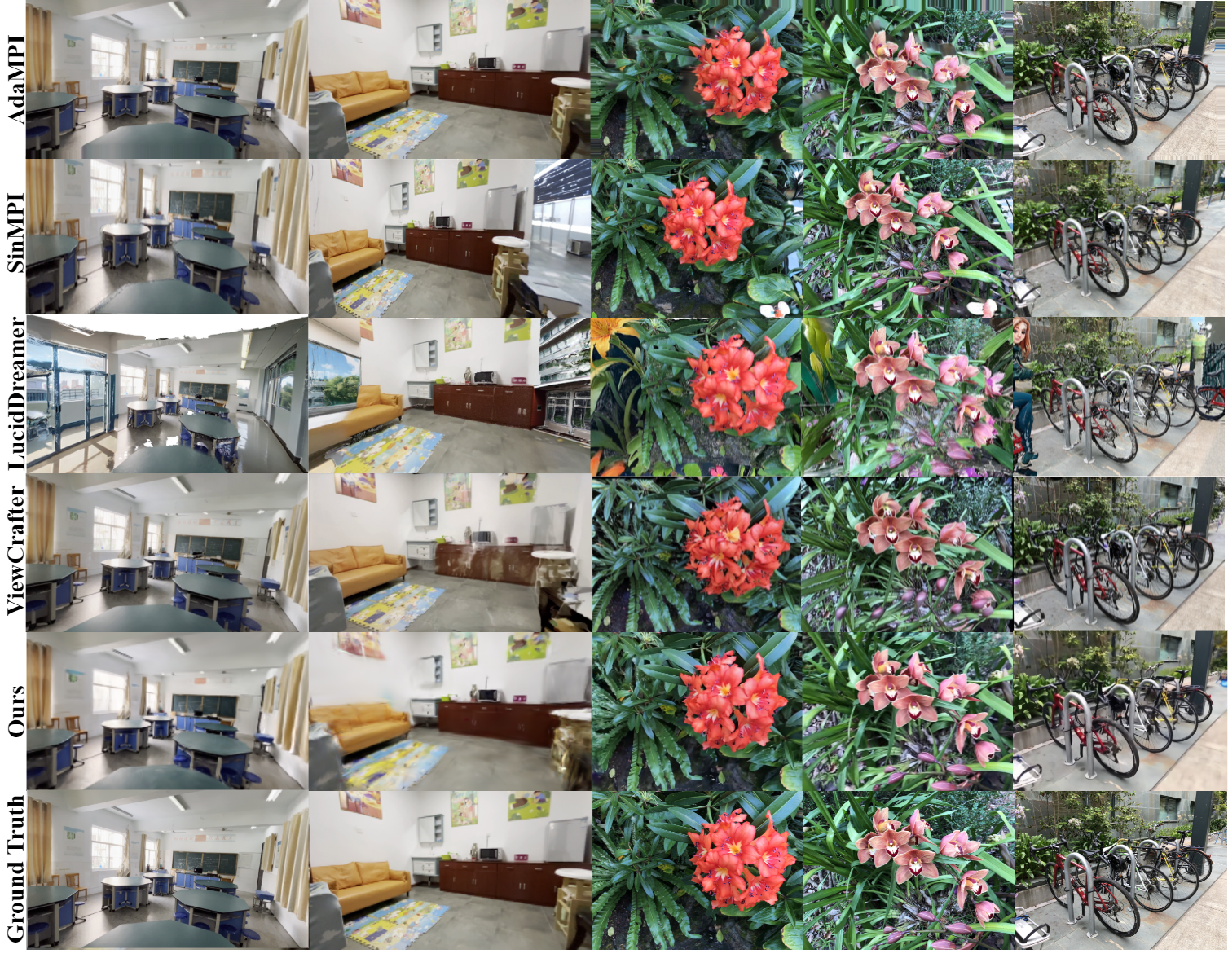}
    \caption{The overall qualitative results of our methods compared with AdaMPI~\cite{Adampi}, SinMPI~\cite{SinMPI}, LucidDreamer~\cite{luciddreamer} and ViewCrafter~\cite{viewcrafter}.
    }
    \label{fig:qualitative}
\end{figure*}
\section{Experiments}
\subsection{Evaluation Protocol and Datasets}
\label{subsec:protocol}
We compare generation quality on the Local Light
Field Fusion (LLFF) dataset~\cite{mildenhall2019llff}, DL3DV dataset~\cite{dl3dv} and Tanks and Temples dataset~\cite{tanksandtemples}. These datasets contain multi-view data that encompass a wide range of complex situations, ranging from indoor to outdoor environments and involving both single and multiple objects, which can be used for evaluating both image quality and 3D consistency. 
For all these datasets~\cite{mildenhall2019llff, dl3dv, tanksandtemples}, the images and the SfM results from COLMAP are provided. We follow the evaluation protocol as AdaMPI~\cite{Adampi} and SinMPI~\cite{SinMPI}, which use a single image as input view and use several surrounding views as ground truth images for quantitative evaluation. Noticing that SinMPI~\cite{SinMPI} and AdaMPI~\cite{Adampi} were only tested on a few samples of LLFF, we choose much more samples to test our model's performance in various and complex scenarios. We randomly selected a total of 20 LLFF~\cite{mildenhall2019llff} scenes, 20 DL3DV~\cite{dl3dv} scenes, and all the test sets of Tanks and Temples dataset~\cite{tanksandtemples} to conduct the aforementioned evaluation protocol.

\subsection{Implementation Details}
We set the number of potential camera pose move directions $D$ to be 4, which contains up, down, left, and right. Following the protocol of MotionCtrl~\cite{wang2023motionctrl}, the LVDM produces 16 frames at a time. The \dustthreer~\cite{mast3r} takes 400 iterations for global alignment. The 3DGS is first trained with 3k iterations in a vanilla setting and then trained with distortion-field network with 14k iterations. 
For the evaluation, all the camera poses and trajectories are estimated via \dustthreer, which may not be well aligned with the ground truth. Therefore, we maintain the 3D Gaussians model trained on training views in a frozen state while optimizing the camera poses for evaluation views, following the setting of InstantSplatting~\cite{fan2024instantsplat} and NeRFmm~\cite{wang2021nerfmm}.

This optimization process focuses on minimizing the photometric discrepancies between the synthesized images and the actual test views, aiming to achieve a more precise alignment for fair comparisons.

\subsection{Quantitative and Qualitative comparison}
As shown in Table~\ref{sample-table}, we compare our \name with previous SOTA methods quantitatively. Our model has shown significant improvements across multiple metrics. In particular, in terms of PSNR, we achieved a 4.73 increase in LLFF scenes and a 3.92 increase in complex outdoor scenes of DL3DV scenes. 

We further compare generation quality on the Tanks and Temples dataset~\cite{tanksandtemples}. Though facing more difficult cases of Tanks and Temples, LiftImage3D still achieves strong performance on Tanks and Temples dataset. Compared to Viewcrafter~\cite{viewcrafter}, LiftImage3D respectively obtained a 2.69 PSNR improvement.

\begin{figure*}
    \centering
    \includegraphics[width=\linewidth]{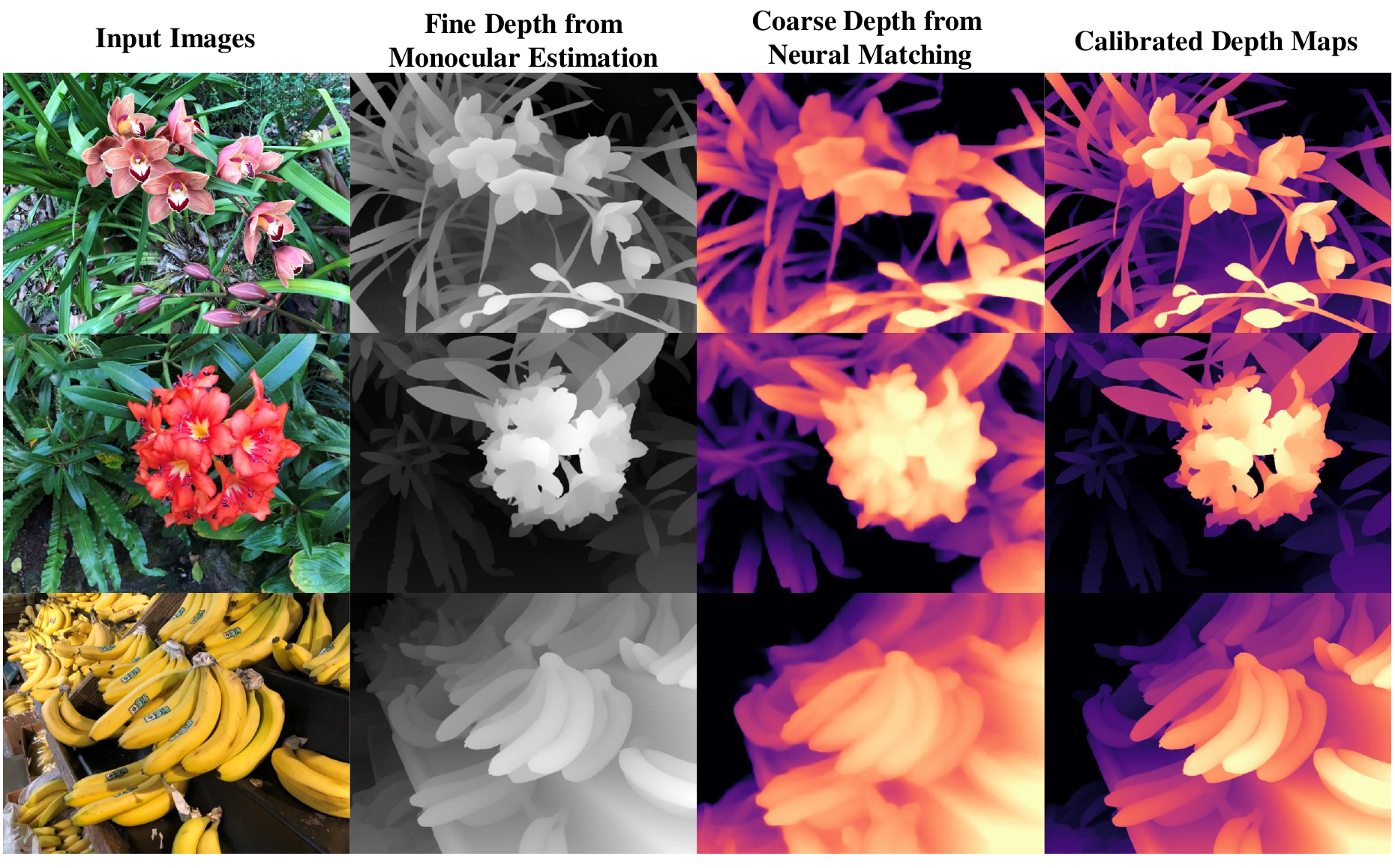}
    \caption{
    Visualization of proposed depth prior injection. The first column lays the video frames generated by LVDM or input images. The second column shows the monocular depth derived from Depth Anything v2~\cite{depth_anything_v2}. The third column shows the coarse depth maps with scale predicted by \dustthreer. The fourth column is the calibrated result providing the fine depth estimates with scales, showing the effectiveness of our depth prior injection module in providing accurate and fine-detailed depth priors. 
    }
    \label{fig:depthalign}
\end{figure*}

As shown in Figure 4., our model exhibits stronger generalization abilities in non-surrounding viewpoints. All displayed images are taken from views that are widely different from the input image. In the case of SinMPI~\cite{SinMPI}, AdaMPI~\cite{Adampi} and LucidDreamer~\cite{luciddreamer}, they tend to make arbitrary estimations for regions far from the surrounding areas. Whether it is through depth guidance or text-to-image diffusion guidance, these models tend to deviate significantly from the input when dealing with scenes that are far from the surrounding context. 
ViewCrafter~\cite{viewcrafter} can produce good visual results, but due to the inaccuracy camera control issues and 3D-inconsistent distortions of video diffusion models, it is challenging for it to match the ground truth. In addition, for complex scenes such as various flowers, the output of ViewCrafter has many detailed distortions.
On the other hand, our model takes full advantage of the prior capabilities of video diffusion, resulting in better 3D consistency. 

\subsection{LiftImage3D with Different LVDMs}
LiftImage3D leverages LVDM’s (MotionCtrl~\cite{wang2023motionctrl} in the manuscript, denoted as LiftImage3D-MotionCtrl) generative priors to achieve high-fidelity single image to 3D reconstruction.  LiftImage3D can also equip with ViewCrafter~\cite{viewcrafter} (LiftImage3D-ViewCrafter) to show its effectiveness across different LVDM priors. 
Specifically, the same as LiftImage3D-MotionCtrl, we also take a set of articulated trajectories as input to Viewcrafter. In contrast to MotionCtrl, the input trajectories are modeled by warping the point maps rather than directly providing camera poses. After obtaining all video frames, we continue to use \dustthreer for neural matching across all video frames. The video frames with camera poses will undergo distortion-aware 3DGS learning based on the poses. Following the setting of ViewCrafter~\cite{viewcrafter}, the LVDM produces 25 frames at a time.
All depth prior injection and loss functions are also applied in the same way as the training process of 3DGS.

The results are shown in Table~\ref{tab:differentlvdm}. Either MotionCtrl or ViewCrafter falls short in building a 3D consistent scene from the input image. With LiftImage3D, the performance improves significantly. Using motionctrl\cite{wang2023motionctrl} as the backbone of LVDM, LiftImage3D achieves a PSNR improvement of 4.07, and when using ViewCrafter~\cite{viewcrafter} as the backbone of LVDM, LiftImage3D also achieves a PSNR improvement of 5.50.
This experiment indicates LiftImage3D's strong ability in ensuring 3D consistency during releasing LVDMs’ generation priors.

\begin{table}
\caption{Quantitative comparisons on the DL3DV dataset with different LVDMs}
  \label{tab:differentlvdm}
  \centering
  \small
    \begin{tabular}{l|ccc}
    \hline 
    Method & PSNR $\uparrow$ & SSIM $\uparrow$ & LPIPS $\downarrow$ \\
    \hline 
    MotionCtrl~\cite{wang2023motionctrl} & 17.00 & 0.597 & 0.371 \\
    LiftImg3D-MotionCtrl  & 21.08 & 0.685 & 0.451 \\
    ViewCrafter~\cite{viewcrafter}  & 17.16 & 0.608 & 0.407 \\
    LiftImg3D-ViewCrafter & 22.66 & 0.728 & 0.387 \\
    \hline
    \end{tabular}
\end{table}

\subsection{Ablation Studies}
For the ablation study, we primarily verified the effectiveness of the distortion-aware 3DGS and its two-dimensional stamp $t = \{t_i, t_j\}$ design. 
We also verify the effectiveness of depth guidance and the loss function $L_{distort}$. 
Specifically, as shown in Table.~\ref{ablation-table}, ``w/o distortion-field'' indicates using vanilla 3DGS to build on the video frames, which shows effectiveness on our distortion-aware 3DGS design.  
``w/o $L_{distort}$" refers to not strictly requiring the distortion field to be zero at the input image, which would lead to the distortion field growing chaotically in various directions. 
This experiment shows that without restriction that completely pursuing modeling the distortion can result in poor canonical 3DGS. The results are much worse than using vanilla 3DGS. 
``w/ 1-axis frame stamp'' indicates using only one-dimensional stamp, which means regardless of the direction that video frames move, we only record their temporal difference relative to the input image as $t$. The performance degradation shows the effectiveness of the two-dimensional stamp design.
``w/o depth-align'' experiments the effectiveness of our depth prior injection module.

\begin{table}
\caption{Ablation studies of our method on LLFF datasets}
  \label{ablation-table}
  \centering
  \small
    \begin{tabular}{l|lll}
    \hline 
     & \small{PSNR $\uparrow$} & \small{SSIM $\uparrow$}  & \small{LPIPS $\downarrow$} \\
    \hline 

      \small{Ours} & \textbf{18.24} &	\textbf{0.488} &	0.519	 \\
     \small{w/o distortion-field}  & 14.93& 0.386& 0.559\\
     \small{w/o $L_{distort}$} & 14.09 & 0.375 & 0.608 \\
     \small{w/ 1-axis frame stamp} &16.00 & 0.430 & \textbf{0.509}\\
     \small{w/o depth-align} & 14.95 & 0.400& 0.572\\
     \hline
    \end{tabular}
\end{table}

We also provide visualizations on our proposed depth prior injections to show its effectiveness as in Figure~\ref{fig:depthalign}.
The figure shows how to leverage the coarse depth map (the left part) from neural matching to calibrate the fine depth map (the right part) from monocular depth estimation with proper scales and shifts. It can be observed that the depth obtained from neural matching often lacks sharpened edges compared to monocular estimation, and some background information may be lost. However, the depth details after calibration are more abundant, with a more precise scale, which can help us to incorporate depth priors into 3DGS~\cite{kerbl3Dgaussians} more accurately.

\section{Conclusion}
This paper presents LiftImage3D, a systematic framework designed to harness the 3D generative capabilities of latent video diffusion models (LVDMs) for transforming any single 2D image into 3D Gaussians. The key idea is to tackle the challenges when using LVDMs to produce multi-view frames, \ie collapse with large camera motions, inaccurate camera control, and independent distortion between frames. A series of techniques are proposed, including articulated trajectory frame generation, robust neural matching-based point cloud registration, and distortion-aware 3D Gaussian splating \emph{etc}. We hope the framework and proposed techniques could inspire future research on facilitating 3D reconstruction with video generation priors.

{
    \small
    \bibliographystyle{ieeenat_fullname}
    \bibliography{main}

\begin{thebibliography}{93}
\providecommand{\natexlab}[1]{#1}
\providecommand{\url}[1]{\texttt{#1}}
\expandafter\ifx\csname urlstyle\endcsname\relax
  \providecommand{\doi}[1]{doi: #1}\else
  \providecommand{\doi}{doi: \begingroup \urlstyle{rm}\Url}\fi

\bibitem[Blattmann et~al.(2023)Blattmann, Dockhorn, Kulal, Mendelevitch, Kilian, and Lorenz]{Blattmann2023svd}
A. Blattmann, T. Dockhorn, S. Kulal, D. Mendelevitch, M. Kilian, and D Lorenz.
\newblock Text-to-3d using gaussian splatting.
\newblock \emph{arXiv:2311.15127}, 2023.

\bibitem[Cao and Johnson(2023)]{cao2023hexplane}
Ang Cao and Justin Johnson.
\newblock Hexplane: A fast representation for dynamic scenes.
\newblock In \emph{Proceedings of the IEEE/CVF Conference on Computer Vision and Pattern Recognition}, pages 130--141, 2023.

\bibitem[Chan et~al.(2023)Chan, Nagano, Chan, Bergman, Park, Levy, Aittala, Mello, Karras, and Wetzstein]{chan2023genvs}
Eric~R. Chan, Koki Nagano, Matthew~A. Chan, Alexander~W. Bergman, Jeong~Joon Park, Axel Levy, Miika Aittala, Shalini~De Mello, Tero Karras, and Gordon Wetzstein.
\newblock {GeNVS}: Generative novel view synthesis with {3D}-aware diffusion models.
\newblock In \emph{arXiv}, 2023.

\bibitem[Chaurasia et~al.(2020)Chaurasia, Nieuwoudt, Ichim, Szeliski, and Sorkine-Hornung]{chaurasia2020}
Gaurav Chaurasia, Arthur Nieuwoudt, Alexandru-Eugen Ichim, Richard Szeliski, and Alexander Sorkine-Hornung.
\newblock Passthrough+ real-time stereoscopic view synthesis for mobile mixed reality.
\newblock \emph{Proceedings of the ACM on Computer Graphics and Interactive Techniques}, 3\penalty0 (1):\penalty0 1--17, 2020.

\bibitem[Chen et~al.(2023)Chen, Ni, Jiang, Zhang, Zhu, and Huang]{chen2023singleview}
Yixin Chen, Junfeng Ni, Nan Jiang, Yaowei Zhang, Yixin Zhu, and Siyuan Huang.
\newblock Single-view 3d scene reconstruction with high-fidelity shape and texture.
\newblock \emph{arXiv:2311.00457}, 2023.

\bibitem[Choi et~al.(2019)Choi, Gallo, Troccoli, Kim, and Kautz]{choi2019}
Inchang Choi, Orazio Gallo, Alejandro Troccoli, Min~H Kim, and Jan Kautz.
\newblock Extreme view synthesis.
\newblock In \emph{Proceedings of the IEEE/CVF International Conference on Computer Vision}, pages 7781--7790, 2019.

\bibitem[Chung et~al.(2023)Chung, Lee, Nam, Lee, and Lee]{luciddreamer}
Jaeyoung Chung, Suyoung Lee, Hyeongjin Nam, Jaerin Lee, and Kyoung~Mu Lee.
\newblock Luciddreamer: Domain-free generation of 3d gaussian splatting scenes.
\newblock \emph{arXiv preprint arXiv:2311.13384}, 2023.

\bibitem[Deitke et~al.(2023{\natexlab{a}})Deitke, Liu, Wallingford, Ngo, Michel, Kusupati, Fan, Laforte, Voleti, Gadre, VanderBilt, Kembhavi, Vondrick, Gkioxari, Ehsani, Schmidt, and Farhadi]{objaverseXL}
Matt Deitke, Ruoshi Liu, Matthew Wallingford, Huong Ngo, Oscar Michel, Aditya Kusupati, Alan Fan, Christian Laforte, Vikram Voleti, Samir~Yitzhak Gadre, Eli VanderBilt, Aniruddha Kembhavi, Carl Vondrick, Georgia Gkioxari, Kiana Ehsani, Ludwig Schmidt, and Ali Farhadi.
\newblock Objaverse-xl: A universe of 10m+ 3d objects.
\newblock \emph{arXiv preprint arXiv:2307.05663}, 2023{\natexlab{a}}.

\bibitem[Deitke et~al.(2023{\natexlab{b}})Deitke, Schwenk, Salvador, Weihs, Michel, VanderBilt, Schmidt, Ehsani, Kembhavi, and Farhadi]{deitke2023objaverse}
Matt Deitke, Dustin Schwenk, Jordi Salvador, Luca Weihs, Oscar Michel, Eli VanderBilt, Ludwig Schmidt, Kiana Ehsani, Aniruddha Kembhavi, and Ali Farhadi.
\newblock Objaverse: A universe of annotated 3d objects.
\newblock In \emph{CVPR}, pages 13142--13153, 2023{\natexlab{b}}.

\bibitem[Fan et~al.(2024)Fan, Cong, Wen, Wang, Zhang, Ding, Xu, Ivanovic, Pavone, Pavlakos, Wang, and Wang]{fan2024instantsplat}
Zhiwen Fan, Wenyan Cong, Kairun Wen, Kevin Wang, Jian Zhang, Xinghao Ding, Danfei Xu, Boris Ivanovic, Marco Pavone, Georgios Pavlakos, Zhangyang Wang, and Yue Wang.
\newblock Instantsplat: Unbounded sparse-view pose-free gaussian splatting in 40 seconds, 2024.

\bibitem[Fridovich-Keil et~al.(2023)Fridovich-Keil, Meanti, Warburg, Recht, and Kanazawa]{kplanes_2023}
Sara Fridovich-Keil, Giacomo Meanti, Frederik~Rahbæk Warburg, Benjamin Recht, and Angjoo Kanazawa.
\newblock K-planes: Explicit radiance fields in space, time, and appearance.
\newblock In \emph{CVPR}, 2023.

\bibitem[Guo et~al.(2023{\natexlab{a}})Guo, Wang, and Lian]{SinMPI}
Pu Guo, Peng-Shuai Wang, and Zhouhui Lian.
\newblock {SinMPI}: Novel view synthesis from a single image with expanded multiplane images.
\newblock \emph{ACM Transactions on Graphics (SIGGRAPH Asia)}, 42\penalty0 (6), 2023{\natexlab{a}}.

\bibitem[Guo et~al.(2023{\natexlab{b}})Guo, Wang, and Lian]{guo2023SinMPI}
Pu Guo, Peng-Shuai Wang, and Zhouhui Lian.
\newblock {SinMPI}: Novel view synthesis from a single image with expanded multiplane images.
\newblock In \emph{SIGGRAPH Asia 2023 Conference Papers}, pages 1--9. ACM, 2023{\natexlab{b}}.

\bibitem[Hamdi et~al.(2023)Hamdi, Ghanem, and Nie{\ss}sner]{hamdi2023sparf}
Abdullah Hamdi, Bernard Ghanem, and Matthias Nie{\ss}sner.
\newblock Sparf: Large-scale learning of 3d sparse radiance fields from few input images.
\newblock In \emph{ICCV}, pages 2930--2940, 2023.

\bibitem[Han et~al.(2024)Han, Kokkinos, and Torr]{Han2024vfusion3d}
Junlin Han, Filippos Kokkinos, and Philip Torr.
\newblock {VFusion3D}: Learning scalable {3D} generative models from video diffusion models.
\newblock \emph{arXiv preprint arXiv:2403.12034}, 2024.

\bibitem[Han et~al.(2022)Han, Wang, and Yang]{Adampi}
Yuxuan Han, Ruicheng Wang, and Jiaolong Yang.
\newblock Single-view view synthesis in the wild with learned adaptive multiplane images.
\newblock In \emph{ACM SIGGRAPH}, 2022.

\bibitem[Hu et~al.(2023)Hu, Hong, Hu, Pan, Mei, Xiao, Yang, and Liu]{hu2023humanliff}
Shoukang Hu, Fangzhou Hong, Tao Hu, Liang Pan, Haiyi Mei, Weiye Xiao, Lei Yang, and Ziwei Liu.
\newblock Humanliff: Layer-wise 3d human generation with diffusion model.
\newblock \emph{arXiv:2308.09712}, 2023.

\bibitem[Huang et~al.(2024)Huang, Stojanov, Thai, Jampani, and Rehg]{huang2024zeroshape}
Zixuan Huang, Stefan Stojanov, Anh Thai, Varun Jampani, and James~M. Rehg.
\newblock Zeroshape: Regression-based zero-shot shape reconstruction.
\newblock \emph{arXiv:2312.14198}, 2024.

\bibitem[Jun and Nichol(2023)]{jun2023shapee}
Heewoo Jun and Alex. et~al. Nichol.
\newblock Shap-e: Generating conditional 3d implicit functions.
\newblock \emph{arXiv:2305.02463}, 2023.

\bibitem[Kerbl et~al.(2023)Kerbl, Kopanas, Leimk{\"u}hler, and Drettakis]{kerbl3Dgaussians}
Bernhard Kerbl, Georgios Kopanas, Thomas Leimk{\"u}hler, and George Drettakis.
\newblock 3d gaussian splatting for real-time radiance field rendering.
\newblock \emph{ACM Transactions on Graphics}, 2023.

\bibitem[Khakhulin et~al.(2022)Khakhulin, Korzhenkov, Solovev, Sterkin, Ardelean, and Lempitsky]{khakhulin2022}
Taras Khakhulin, Denis Korzhenkov, Pavel Solovev, Gleb Sterkin, Andrei-Timotei Ardelean, and Victor Lempitsky.
\newblock Stereo magnification with multi-layer images.
\newblock In \emph{Proceedings of the IEEE/CVF Conference on Computer Vision and Pattern Recognition}, pages 8687--8696, 2022.

\bibitem[Khan et~al.(2023)Khan, Numair, Xiao, and Lanman.]{barron2023tiled3dphotography}
Khan, Numair, Lei Xiao, and Douglas Lanman.
\newblock Tiled multiplane images for practical 3d photography.
\newblock In \emph{ICCV}, 2023.

\bibitem[Knapitsch et~al.(2017)Knapitsch, Park, Zhou, and Koltun]{tanksandtemples}
Arno Knapitsch, Jaesik Park, Qian-Yi Zhou, and Vladlen Koltun.
\newblock Tanks and temples: Benchmarking large-scale scene reconstruction.
\newblock \emph{ACM Transactions on Graphics}, 36\penalty0 (4), 2017.

\bibitem[Kocsis et~al.(2023)Kocsis, Sitzmann, and Nießner]{kocsis2023intrinsic}
Peter Kocsis, Vincent Sitzmann, and Matthias Nießner.
\newblock Intrinsic image diffusion for single-view material estimation.
\newblock \emph{arXiv:2312.12274}, 2023.

\bibitem[Kwak et~al.(2023)Kwak, Dong, Jin, Ko, Mahajan, and Yi]{kwak2023vivid123}
Jun-Gyu Kwak, Eunhyeok Dong, Youngho Jin, Hosik Ko, Surbhi Mahajan, and Kwang~Moo Yi.
\newblock {Vivid-1-to-3}: Novel view synthesis with video diffusion models.
\newblock \emph{arXiv preprint arXiv:2312.01305}, 2023.

\bibitem[Lee et~al.(2023)Lee, Kim, Cho, and Han]{lee2023localaware}
D. Lee, C. Kim, M. Cho, and W.~S. Han.
\newblock Locality-aware generalizable implicit neural representation.
\newblock In \emph{arXiv:2310.05624}, 2023.

\bibitem[Leroy et~al.(2024)Leroy, Cabon, and Revaud]{mast3r}
Vincent Leroy, Yohann Cabon, and J{\'e}r{\^o}me Revaud.
\newblock Grounding image matching in 3d with mast3r.
\newblock \emph{arXiv preprint arXiv:2406.09756}, 2024.

\bibitem[Li et~al.(2022)Li, Zhang, H{\"a}ne, Tang, Varshney, and Du]{li2022}
David Li, Yinda Zhang, Christian H{\"a}ne, Danhang Tang, Amitabh Varshney, and Ruofei Du.
\newblock Omnisyn: Synthesizing 360 videos with wide-baseline panoramas.
\newblock In \emph{2022 IEEE Conference on Virtual Reality and 3D User Interfaces Abstracts and Workshops (VRW)}, pages 670--671. IEEE, 2022.

\bibitem[Li and Kalantari(2020)]{li2020}
Qinbo Li and Nima~Khademi Kalantari.
\newblock Synthesizing light field from a single image with variable mpi and two network fusion.
\newblock \emph{ACM Trans. Graph.}, 39\penalty0 (6):\penalty0 229--1, 2020.

\bibitem[Li et~al.(2023)Li, Zanjani, Yahia, Asano, Gall, and Habibian]{li2023valid}
Shijie Li, Farhad~G. Zanjani, Haitam~Ben Yahia, Yuki~M. Asano, Juergen Gall, and Amirhossein Habibian.
\newblock Valid: Variable-length input diffusion for novel view synthesis.
\newblock \emph{arXiv:2312.08892}, 2023.

\bibitem[Lin et~al.(2023)Lin, Han, Gong, Xu, Zhang, and Li]{lin2023consistent123}
Yukang Lin, Haonan Han, Chaoqun Gong, Zunnan Xu, Yachao Zhang, and Xiu Li.
\newblock Consistent123: One image to highly consistent 3d asset using case-aware diffusion priors.
\newblock \emph{arXiv:2309.17261}, 2023.

\bibitem[Ling et~al.(2024)Ling, Sheng, Tu, Zhao, Xin, Wan, Yu, Guo, Yu, Lu, et~al.]{dl3dv}
Lu Ling, Yichen Sheng, Zhi Tu, Wentian Zhao, Cheng Xin, Kun Wan, Lantao Yu, Qianyu Guo, Zixun Yu, Yawen Lu, et~al.
\newblock Dl3dv-10k: A large-scale scene dataset for deep learning-based 3d vision.
\newblock In \emph{Proceedings of the IEEE/CVF Conference on Computer Vision and Pattern Recognition}, pages 22160--22169, 2024.

\bibitem[Liu et~al.(2023{\natexlab{a}})Liu, Shi, Chen, Zhang, Xu, Wei, Chen, Zeng, Gu, and Su]{liu2023one2345}
Minghua Liu, Ruoxi Shi, Linghao Chen, Zhuoyang Zhang, Chao Xu, Xinyue Wei, Hansheng Chen, Chong Zeng, Jiayuan Gu, and Hao Su.
\newblock One-2-3-45++: Fast single image to 3d objects with consistent multi-view generation and 3d diffusion.
\newblock \emph{arXiv:2311.07885}, 2023{\natexlab{a}}.

\bibitem[Liu et~al.(2023{\natexlab{b}})Liu, Xu, Jin, Chen, Xu, Su, et~al.]{liu2023one}
Minghua Liu, Chao Xu, Haian Jin, Linghao Chen, Zexiang Xu, Hao Su, et~al.
\newblock One-2-3-45: Any single image to 3d mesh in 45 seconds without per-shape optimization.
\newblock \emph{arXiv:2306.16928}, 2023{\natexlab{b}}.

\bibitem[Liu et~al.(2023{\natexlab{c}})Liu, Wu, Hoorick, Tokmakov, Zakharov, and Vondrick]{liu2023zero1to3}
Ruoshi Liu, Rundi Wu, Basile~Van Hoorick, Pavel Tokmakov, Sergey Zakharov, and Carl Vondrick.
\newblock Zero-1-to-3: Zero-shot one image to 3d object.
\newblock \emph{arXiv:2303.11328}, 2023{\natexlab{c}}.

\bibitem[Liu et~al.(2023{\natexlab{d}})Liu, Lin, Zeng, Long, Liu, Komura, and Wang]{liu2023syncdreamer}
Yuan Liu, Cheng Lin, Zijiao Zeng, Xiaoxiao Long, Lingjie Liu, Taku Komura, and Wenping Wang.
\newblock Syncdreamer: Generating multiview-consistent images from a single-view image.
\newblock \emph{arXiv:2309.03453}, 2023{\natexlab{d}}.

\bibitem[Liu et~al.(2024)Liu, Zhang, Li, Yan, Gao, and Chen]{liu2024sora}
Y. Liu, K. Zhang, Y. Li, Z. Yan, C. Gao, and R. Chen.
\newblock {Sora}: A review on background, technology, limitations, and opportunities of large vision models.
\newblock \emph{arXiv preprint arXiv:2402.17177}, 2024.

\bibitem[Long et~al.(2023)Long, Guo, Lin, Liu, Dou, Liu, Ma, Zhang, Habermann, Theobalt, et~al.]{long2023wonder3d}
Xiaoxiao Long, Yuan-Chen Guo, Cheng Lin, Yuan Liu, Zhiyang Dou, Lingjie Liu, Yuexin Ma, Song-Hai Zhang, Marc Habermann, Christian Theobalt, et~al.
\newblock Wonder3d: Single image to 3d using cross-domain diffusion.
\newblock \emph{arXiv:2310.15008}, 2023.

\bibitem[Luo and Hancock.(1999)]{Procrustesalignment}
Bin Luo and Edwin~R. Hancock.
\newblock Procrustes alignment with the em algorithm.
\newblock \emph{International Conference on Computer Analysis of Images and Patterns.}, 1999.

\bibitem[Luvizon et~al.(2021)Luvizon, Carvalho, dos Santos, Conceicao, Flores-Campana, Decker, Souza, Pedrini, Joia, and Penatti]{luvizon2021}
Diogo~C Luvizon, Gustavo Sutter~P Carvalho, Andreza~A dos Santos, Jhonatas~S Conceicao, Jose~L Flores-Campana, Luis~GL Decker, Marcos~R Souza, Helio Pedrini, Antonio Joia, and Otavio~AB Penatti.
\newblock Adaptive multiplane image generation from a single internet picture.
\newblock In \emph{Proceedings of the IEEE/CVF Winter Conference on Applications of Computer Vision}, pages 2556--2565, 2021.

\bibitem[Melas-Kyriazi et~al.(2023)Melas-Kyriazi, Laina, Rupprecht, and Vedaldi]{melas2023realfusion}
Luke Melas-Kyriazi, Iro Laina, Christian Rupprecht, and Andrea Vedaldi.
\newblock Realfusion: 360deg reconstruction of any object from a single image.
\newblock In \emph{CVPR}, pages 8446--8455, 2023.

\bibitem[Melas-Kyriazi et~al.(2024)Melas-Kyriazi, Laina, Rupprecht, Neverova, Vedaldi, Gafni, and Kokkinos]{melas2024im3d}
Luke Melas-Kyriazi, Iro Laina, Christian Rupprecht, Natalia Neverova, Andrea Vedaldi, Oran Gafni, and Filippos Kokkinos.
\newblock {IM-3D}: Iterative multiview diffusion and reconstruction for high-quality {3D} generation.
\newblock \emph{arXiv preprint arXiv:2402.08682}, 2024.

\bibitem[Mildenhall et~al.(2019)Mildenhall, Srinivasan, Ortiz-Cayon, Kalantari, Ramamoorthi, Ng, and Kar]{mildenhall2019llff}
Ben Mildenhall, Pratul~P. Srinivasan, Rodrigo Ortiz-Cayon, Nima~Khademi Kalantari, Ravi Ramamoorthi, Ren Ng, and Abhishek Kar.
\newblock Local light field fusion: Practical view synthesis with prescriptive sampling guidelines.
\newblock \emph{ACM Transactions on Graphics (TOG)}, 2019.

\bibitem[Nichol et~al.(2022)Nichol, Jun, Dhariwal, Mishkin, and Chen]{nichol2022point}
Alex Nichol, Heewoo Jun, Prafulla Dhariwal, Pamela Mishkin, and Mark Chen.
\newblock Point-e: A system for generating 3d point clouds from complex prompts.
\newblock \emph{arXiv:2212.08751}, 2022.

\bibitem[Niemeyer et~al.(2021)Niemeyer, Barron, Mildenhall, Sajjadi, Geiger, and Radwan]{niemeyer2021regnerf}
Michael Niemeyer, Jonathan~T Barron, Ben Mildenhall, Mehdi~SM Sajjadi, Andreas Geiger, and Noha Radwan.
\newblock Regnerf: Regularizing neural radiance fields for view synthesis from sparse inputs.
\newblock \emph{arXiv preprint arXiv:2112.00724}, 2021.

\bibitem[Niklaus et~al.(2019)Niklaus, Mai, Yang, and Liu]{niklaus2019}
Simon Niklaus, Long Mai, Jimei Yang, and Feng Liu.
\newblock 3d ken burns effect from a single image.
\newblock \emph{ACM Transactions on Graphics (ToG)}, 38\penalty0 (6):\penalty0 1--15, 2019.

\bibitem[Pang et~al.(2024)Pang, Jia, Shi, Tang, Zhang, and Cheng]{pang2024envision3d}
Yiyu Pang, Tong Jia, Yichun Shi, Zimeng Tang, Jiaxiang Zhang, and Xiaohua Cheng.
\newblock {Envision3D}: One image to {3D} with anchor views interpolation.
\newblock \emph{arXiv preprint arXiv:2403.08902}, 2024.

\bibitem[Plastria(2011)]{Weiszfeldalgorithm}
Frank. Plastria.
\newblock The weiszfeld algorithm: proof, amendments, and extensions.
\newblock \emph{Foundations of location analysis}, 2011.

\bibitem[Qian et~al.(2023)Qian, Mai, Hamdi, Ren, Siarohin, Li, Lee, Skorokhodov, Wonka, Tulyakov, et~al.]{qian2023magic123}
Guocheng Qian, Jinjie Mai, Abdullah Hamdi, Jian Ren, Aliaksandr Siarohin, Bing Li, Hsin-Ying Lee, Ivan Skorokhodov, Peter Wonka, Sergey Tulyakov, et~al.
\newblock Magic123: One image to high-quality 3d object generation using both 2d and 3d diffusion priors.
\newblock \emph{arXiv:2306.17843}, 2023.

\bibitem[Qian et~al.(2024)Qian, Wang, Luo, Zhang, Tai, Zhang, Wang, Xue, Zhao, Huang, Wu, and Fu]{qian2024pushing}
Xuelin Qian, Yu Wang, Simian Luo, Yinda Zhang, Ying Tai, Zhenyu Zhang, Chengjie Wang, Xiangyang Xue, Bo Zhao, Tiejun Huang, Yunsheng Wu, and Yanwei Fu.
\newblock Pushing auto-regressive models for 3d shape generation at capacity and scalability.
\newblock \emph{arXiv:2402.12225}, 2024.

\bibitem[Sargent et~al.(2023)Sargent, Li, Shah, Herrmann, Yu, Zhang, Chan, Lagun, Fei-Fei, Sun, et~al.]{sargent2023zeronvs}
Kyle Sargent, Zizhang Li, Tanmay Shah, Charles Herrmann, Hong-Xing Yu, Yunzhi Zhang, Eric~Ryan Chan, Dmitry Lagun, Li Fei-Fei, Deqing Sun, et~al.
\newblock Zeronvs: Zero-shot 360-degree view synthesis from a single real image.
\newblock \emph{arXiv:2310.17994}, 2023.

\bibitem[Sch\"{o}nberger and Frahm(2016)]{schoenberger2016sfm}
Johannes~Lutz Sch\"{o}nberger and Jan-Michael Frahm.
\newblock Structure-from-motion revisited.
\newblock In \emph{Conference on Computer Vision and Pattern Recognition (CVPR)}, 2016.

\bibitem[Sch\"{o}nberger et~al.(2016)Sch\"{o}nberger, Zheng, Pollefeys, and Frahm]{schoenberger2016mvs}
Johannes~Lutz Sch\"{o}nberger, Enliang Zheng, Marc Pollefeys, and Jan-Michael Frahm.
\newblock Pixelwise view selection for unstructured multi-view stereo.
\newblock In \emph{European Conference on Computer Vision (ECCV)}, 2016.

\bibitem[Shen et~al.(2023)Shen, Yang, and Wang]{shen2023anything}
Qiuhong Shen, Xingyi Yang, and Xinchao Wang.
\newblock Anything-3d: Towards single-view anything reconstruction in the wild.
\newblock \emph{arXiv:2304.10261}, 2023.

\bibitem[Shi et~al.(2023{\natexlab{a}})Shi, Chen, Zhang, Liu, Xu, Wei, Chen, Zeng, and Su]{shi2023zero123++}
Ruoxi Shi, Hansheng Chen, Zhuoyang Zhang, Minghua Liu, Chao Xu, Xinyue Wei, Linghao Chen, Chong Zeng, and Hao Su.
\newblock Zero123++: a single image to consistent multi-view diffusion base model.
\newblock \emph{arXiv:2310.15110}, 2023{\natexlab{a}}.

\bibitem[Shi et~al.(2023{\natexlab{b}})Shi, Wang, Cao, Tang, Qi, Yang, Huang, Liu, Zhang, and Shum]{shi2023toss}
Yukai Shi, Jianan Wang, He Cao, Boshi Tang, Xianbiao Qi, Tianyu Yang, Yukun Huang, Shilong Liu, Lei Zhang, and Heung-Yeung Shum.
\newblock Toss: High-quality text-guided novel view synthesis from a single image.
\newblock \emph{arXiv:2310.10644}, 2023{\natexlab{b}}.

\bibitem[Shi et~al.(2023{\natexlab{c}})Shi, Wang, Cao, Tang, Qi, Yang, Huang, Liu, Zhang, and Shum]{shi2023tosshighquality}
Yukai Shi, Jianan Wang, He Cao, Boshi Tang, Xianbiao Qi, Tianyu Yang, Yukun Huang, Shilong Liu, Lei Zhang, and Heung-Yeung Shum.
\newblock Toss:high-quality text-guided novel view synthesis from a single image.
\newblock \emph{arXiv:2310.10644}, 2023{\natexlab{c}}.

\bibitem[Shih et~al.(2020)Shih, Su, Kopf, and Huang]{shih20}
Meng-Li Shih, Shih-Yang Su, Johannes Kopf, and Jia-Bin Huang.
\newblock 3d photography using context-aware layered depth inpainting.
\newblock In \emph{IEEE Conference on Computer Vision and Pattern Recognition (CVPR)}, 2020.

\bibitem[Simon et~al.(2024)Simon, He, Perez-Rua, Xu, Benhalloum, and Xiang]{simon2024hypervoltran}
Christian Simon, Sen He, Juan-Manuel Perez-Rua, Mengmeng Xu, Amine Benhalloum, and Tao Xiang.
\newblock Hyper-voltran: Fast and generalizable one-shot image to 3d object structure via hypernetworks.
\newblock \emph{arXiv:2312.16218}, 2024.

\bibitem[Srinivasan et~al.(2019)Srinivasan, Tucker, Barron, Ramamoorthi, Ng, and Snavely]{srinivasan2019}
Pratul~P Srinivasan, Richard Tucker, Jonathan~T Barron, Ravi Ramamoorthi, Ren Ng, and Noah Snavely.
\newblock Pushing the boundaries of view extrapolation with multiplane images.
\newblock In \emph{Proceedings of the IEEE/CVF Conference on Computer Vision and Pattern Recognition}, pages 175--184, 2019.

\bibitem[Szeliski and Golland(1998)]{szeliski1998}
Richard Szeliski and Polina Golland.
\newblock Stereo matching with transparency and matting.
\newblock In \emph{Sixth International Conference on Computer Vision (IEEE Cat. No. 98CH36271)}, pages 517--524. IEEE, 1998.

\bibitem[Tang et~al.(2023)Tang, Wang, Zhang, Zhang, Yi, Ma, and Chen]{tang2023make}
Junshu Tang, Tengfei Wang, Bo Zhang, Ting Zhang, Ran Yi, Lizhuang Ma, and Dong Chen.
\newblock Make-it-3d: High-fidelity 3d creation from a single image with diffusion prior.
\newblock \emph{arXiv:2303.14184}, 2023.

\bibitem[Tang et~al.(2024)Tang, Chen, Wang, Tang, Zhang, Fan, Chandra, Furukawa, and Ranjan]{tang2024mvdiffusion}
Shitao Tang, Jiacheng Chen, Dilin Wang, Chengzhou Tang, Fuyang Zhang, Yuchen Fan, Vikas Chandra, Yasutaka Furukawa, and Rakesh Ranjan.
\newblock Mvdiffusion++: A dense high-resolution multi-view diffusion model for single or sparse-view 3d object reconstruction.
\newblock \emph{arXiv:2402.12712}, 2024.

\bibitem[Voleti et~al.(2024)Voleti, Yao, Boss, Letts, Pankratz, and Tochilkin]{voleti2024sv3d}
Vikram Voleti, Cheng-Hao Yao, Matthew Boss, Alexander Letts, Daniel Pankratz, and Denis Tochilkin.
\newblock {SV3D}: Novel multi-view synthesis and {3D} generation from a single image using latent video diffusion.
\newblock \emph{arXiv preprint arXiv:2403.12008}, 2024.

\bibitem[Wang et~al.(2024)Wang, Leroy, Cabon, Chidlovskii, and Revaud]{dust3r_cvpr24}
Shuzhe Wang, Vincent Leroy, Yohann Cabon, Boris Chidlovskii, and Jerome Revaud.
\newblock Dust3r: Geometric 3d vision made easy.
\newblock In \emph{CVPR}, 2024.

\bibitem[Wang et~al.(2021)Wang, Wu, Xie, Chen, and Prisacariu]{wang2021nerfmm}
Zirui Wang, Shangzhe Wu, Weidi Xie, Min Chen, and Victor~Adrian Prisacariu.
\newblock Ne{RF}$--$: Neural radiance fields without known camera parameters.
\newblock \emph{arXiv preprint arXiv:2102.07064}, 2021.

\bibitem[Wang et~al.(2023)Wang, Yuan, Wang, Chen, Xia, Luo, and Shan]{wang2023motionctrl}
Zhouxia Wang, Ziyang Yuan, Xintao Wang, Tianshui Chen, Menghan Xia, Ping Luo, and Yin Shan.
\newblock Motionctrl: A unified and flexible motion controller for video generation.
\newblock In \emph{arXiv preprint arXiv:2312.03641}, 2023.

\bibitem[Weng et~al.(2023{\natexlab{a}})Weng, Yang, Wang, Li, Zhang, Chen, and Zhang]{weng2023consistent123}
Haohan Weng, Tianyu Yang, Jianan Wang, Yu Li, Tong Zhang, CL Chen, and Lei Zhang.
\newblock Consistent123: Improve consistency for one image to 3d object synthesis.
\newblock \emph{arXiv:2310.08092}, 2023{\natexlab{a}}.

\bibitem[Weng et~al.(2023{\natexlab{b}})Weng, Wang, and Yeung]{weng2023zeroavatar}
Zhenzhen Weng, Zeyu Wang, and Serena Yeung.
\newblock Zeroavatar: Zero-shot 3d avatar generation from a single image.
\newblock \emph{arXiv:2305.16411}, 2023{\natexlab{b}}.

\bibitem[Wiles et~al.(2020)Wiles, Gkioxari, Szeliski, and Johnson]{wiles2020}
Olivia Wiles, Georgia Gkioxari, Richard Szeliski, and Justin Johnson.
\newblock Synsin: End-to-end view synthesis from a single image.
\newblock In \emph{Proceedings of the IEEE/CVF Conference on Computer Vision and Pattern Recognition}, pages 7467--7477, 2020.

\bibitem[with Stunning Geometrical~Consistency.(2024)]{li2024sora}
Sora Generates~Videos with Stunning Geometrical~Consistency.
\newblock Li, x., zhou, d., zhang, c., wei, s., hou, q., and cheng, m. m.
\newblock \emph{arXiv:2402.17403}, 2024.

\bibitem[Wizadwongsa et~al.(2021)Wizadwongsa, Phongthawee, Yenphraphai, and Suwajanakorn]{wizadwongsa2021}
Suttisak Wizadwongsa, Pakkapon Phongthawee, Jiraphon Yenphraphai, and Supasorn Suwajanakorn.
\newblock Nex: Real-time view synthesis with neural basis expansion.
\newblock In \emph{Proceedings of the IEEE/CVF Conference on Computer Vision and Pattern Recognition}, pages 8534--8543, 2021.

\bibitem[Woo et~al.(2023)Woo, Park, Go, Kim, and Kim]{woo2023harmonyview}
Sangmin Woo, Byeongjun Park, Hyojun Go, Jin-Young Kim, and Changick Kim.
\newblock Harmonyview: Harmonizing consistency and diversity in one-image-to-3d.
\newblock \emph{arXiv:2312.15980}, 2023.

\bibitem[Wu et~al.(2023{\natexlab{a}})Wu, Yi, Fang, Xie, Zhang, Wei, Liu, Tian, and Xinggang]{wu20234dgaussians}
Guanjun Wu, Taoran Yi, Jiemin Fang, Lingxi Xie, Xiaopeng Zhang, Wei Wei, Wenyu Liu, Qi Tian, and Wang Xinggang.
\newblock 4d gaussian splatting for real-time dynamic scene rendering.
\newblock \emph{arXiv preprint arXiv:2310.08528}, 2023{\natexlab{a}}.

\bibitem[Wu et~al.(2023{\natexlab{b}})Wu, Li, Yang, Zhang, Pan, Wang, Lin, and Liu]{wu2023hyperdreamer}
Tong Wu, Zhibing Li, Shuai Yang, Pan Zhang, Xingang Pan, Jiaqi Wang, Dahua Lin, and Ziwei Liu.
\newblock Hyperdreamer: Hyper-realistic 3d content generation and editing from a single image.
\newblock In \emph{SIGGRAPH Asia 2023 Conference Papers}, 2023{\natexlab{b}}.

\bibitem[Xiang et~al.(2023)Xiang, Yang, Huang, and Tong]{xiang20233daware}
Jianfeng Xiang, Jiaolong Yang, Binbin Huang, and Xin Tong.
\newblock 3d-aware image generation using 2d diffusion models.
\newblock In \emph{Proceedings of the IEEE/CVF International Conference on Computer Vision (ICCV)}, pages 2383--2393, 2023.

\bibitem[Xiao et~al.(2022)Xiao, Nouri, Hegland, Garcia, and Lanman]{xiao2022}
Lei Xiao, Salah Nouri, Joel Hegland, Alberto~Garcia Garcia, and Douglas Lanman.
\newblock Neuralpassthrough: Learned real-time view synthesis for vr.
\newblock In \emph{ACM SIGGRAPH 2022 Conference Proceedings}, pages 1--9, 2022.

\bibitem[Xu et~al.(2022)Xu, Jiang, Wang, Fan, Shi, and Wang]{Xu_2022_SinNeRF}
Dejia Xu, Yifan Jiang, Peihao Wang, Zhiwen Fan, Humphrey Shi, and Zhangyang Wang.
\newblock Sinnerf: Training neural radiance fields on complex scenes from a single image.
\newblock 2022.

\bibitem[Xu et~al.(2023)Xu, Jiang, Wang, Fan, Wang, and Wang]{xu2023neurallift}
Dejia Xu, Yifan Jiang, Peihao Wang, Zhiwen Fan, Yi Wang, and Zhangyang Wang.
\newblock Neurallift-360: Lifting an in-the-wild 2d photo to a 3d object with 360deg views.
\newblock In \emph{CVPR}, pages 4479--4489, 2023.

\bibitem[Yang et~al.(2024{\natexlab{a}})Yang, Li, Fang, Liang, Xie, Zhang, Shen, and Tian]{yang2024gaussianobject}
Chen Yang, Sikuang Li, Jiemin Fang, Ruofan Liang, Lingxi Xie, Xiaopeng Zhang, Wei Shen, and Qi Tian.
\newblock Gaussianobject: Just taking four images to get a high-quality 3d object with gaussian splatting.
\newblock \emph{arXiv:2402.10259}, 2024{\natexlab{a}}.

\bibitem[Yang et~al.(2023{\natexlab{a}})Yang, Cheng, Duan, Ji, and Li]{yang2023consistnet}
Jiayu Yang, Ziang Cheng, Yunfei Duan, Pan Ji, and Hongdong Li.
\newblock Consistnet: Enforcing 3d consistency for multi-view images diffusion.
\newblock \emph{arXiv:2310.10343}, 2023{\natexlab{a}}.

\bibitem[Yang et~al.(2024{\natexlab{b}})Yang, Kang, Huang, Zhao, Xu, Feng, and Zhao]{depth_anything_v2}
Lihe Yang, Bingyi Kang, Zilong Huang, Zhen Zhao, Xiaogang Xu, Jiashi Feng, and Hengshuang Zhao.
\newblock Depth anything v2.
\newblock \emph{arXiv:2406.09414}, 2024{\natexlab{b}}.

\bibitem[Yang et~al.(2023{\natexlab{b}})Yang, Gao, Zhou, Jiao, Zhang, and Jin]{yang2023deformable3dgs}
Ziyi Yang, Xinyu Gao, Wen Zhou, Shaohui Jiao, Yuqing Zhang, and Xiaogang Jin.
\newblock Deformable 3d gaussians for high-fidelity monocular dynamic scene reconstruction.
\newblock \emph{arXiv preprint arXiv:2309.13101}, 2023{\natexlab{b}}.

\bibitem[Ye et~al.(2024)Ye, Liu, Xu, Xueting, Pollefeys, Yang, and Songyou]{ye2024noposplat}
Botao Ye, Sifei Liu, Haofei Xu, Li Xueting, Marc Pollefeys, Ming-Hsuan Yang, and Peng Songyou.
\newblock No pose, no problem: Surprisingly simple 3d gaussian splats from sparse unposed images.
\newblock \emph{arXiv preprint arXiv:2410.24207}, 2024.

\bibitem[Ye et~al.(2023)Ye, Wang, Li, Shi, and Wang]{ye2023consistent}
Jianglong Ye, Peng Wang, Kejie Li, Yichun Shi, and Heng Wang.
\newblock Consistent-1-to-3: Consistent image to 3d view synthesis via geometry-aware diffusion models.
\newblock \emph{arXiv:2310.03020}, 2023.

\bibitem[You et~al.(2024)You, Zhu, Liu, and Hou]{nvssolver}
Meng You, Zhiyu Zhu, Hui Liu, and Junhui Hou.
\newblock Nvs-solver: Video diffusion model as zero-shot novel view synthesizer.
\newblock \emph{arXiv preprint arXiv:2405.15364}, 2024.

\bibitem[Yu et~al.(2024{\natexlab{a}})Yu, Duan, Herrmann, Freeman, and Wu]{wonderworld}
Hong-Xing Yu, Haoyi Duan, Charles Herrmann, William~T Freeman, and Jiajun Wu.
\newblock Wonderworld: Interactive 3d scene generation from a single image.
\newblock \emph{arXiv preprint arXiv:2406.09394}, 2024{\natexlab{a}}.

\bibitem[Yu et~al.(2023)Yu, Liu, Feng, Cui, and Xie]{yu2023boosting3d}
Kai Yu, Jinlin Liu, Mengyang Feng, Miaomiao Cui, and Xuansong Xie.
\newblock Boosting3d: High-fidelity image-to-3d by boosting 2d diffusion prior to 3d prior with progressive learning.
\newblock \emph{arXiv:2311.13617}, 2023.

\bibitem[Yu et~al.(2024{\natexlab{b}})Yu, Xing, Yuan, Hu, Li, Huang, Gao, Wong, Shan, and Tian]{viewcrafter}
Wangbo Yu, Jinbo Xing, Li Yuan, Wenbo Hu, Xiaoyu Li, Zhipeng Huang, Xiangjun Gao, Tien-Tsin Wong, Ying Shan, and Yonghong Tian.
\newblock Viewcrafter: Taming video diffusion models for high-fidelity novel view synthesis.
\newblock \emph{arXiv preprint arXiv:2409.02048}, 2024{\natexlab{b}}.

\bibitem[Zhang et~al.(2023)Zhang, Tang, Pang, Cheng, Jin, Wei, Ning, and Yuan]{zhang2023repaint123}
Junwu Zhang, Zhenyu Tang, Yatian Pang, Xinhua Cheng, Peng Jin, Yida Wei, Munan Ning, and Li Yuan.
\newblock Repaint123: Fast and high-quality one image to 3d generation with progressive controllable 2d repainting.
\newblock \emph{arXiv:2312.13271}, 2023.

\bibitem[Zhang et~al.(2024)Zhang, Herrmann, Hur, Jampani, Darrell, Cole, Sun, and Yang]{monst3r}
Junyi Zhang, Charles Herrmann, Junhwa Hur, Varun Jampani, Trevor Darrell, Forrester Cole, Deqing Sun, and Ming-Hsuan Yang.
\newblock Monst3r: A simple approach for estimating geometry in the presence of motion.
\newblock \emph{arXiv preprint arXiv:2410.03825}, 2024.

\bibitem[Zhang et~al.(2018)Zhang, Isola, Efros, Shechtman, and Wang]{lpipsloss}
Richard Zhang, Phillip Isola, Alexei~A Efros, Eli Shechtman, and Oliver Wang.
\newblock The unreasonable effectiveness of deep features as a perceptual metric.
\newblock In \emph{CVPR}, 2018.

\bibitem[Zhou et~al.(2018)Zhou, Tucker, Flynn, Fyffe, and Snavely]{zhou2018}
Tinghui Zhou, Richard Tucker, John Flynn, Graham Fyffe, and Noah Snavely.
\newblock Stereo magnification: Learning view synthesis using multiplane images.
\newblock In \emph{SIGGRAPH}, 2018.

\end{thebibliography}
}

\end{document}